\documentclass[]{spie}  

\usepackage{amsfonts}
\usepackage{amsmath}
\usepackage{amsmath,bm} 
 
\usepackage{amsmath,amsfonts,amssymb}
\usepackage{graphicx}
\usepackage[colorlinks=true, allcolors=blue]{hyperref}
\usepackage{xcolor, soul}

\title{Treatment-wise Glioblastoma Survival Inference with Multi-parametric Preoperative MRI}

\author[a]{Xiaofeng Liu}
\author[b]{Nadya Shusharina}
\author[b]{Helen A Shih}
\author[c]{C.-C. Jay Kuo}
\author[d]{Georges El Fakhri}
\author[a]{Jonghye Woo}

\affil[a]{Gordon Center for Medical Imaging, Massachusetts
General Hospital and Harvard Medical School, Boston, MA 02114 USA}
\affil[b]{Dept. of Radiation Oncology, Massachusetts General Hospital and Harvard Medical School, Boston, MA 02114, USA}
\affil[c]{Dept. of Electrical and Computer Engineering, University of Southern California, Los Angeles, CA 90007, USA}
\affil[d]{Dept. of Radiology and Biomedical Imaging, Yale University, New Heaven, CT 06519, USA}

\begin{document} 
\maketitle

\begin{abstract}


In this work, we aim to predict the survival time (ST) of glioblastoma (GBM) patients undergoing different treatments based on preoperative magnetic resonance (MR) scans. The personalized and precise treatment planning can be achieved by comparing the ST of different treatments. It is well established that both the current status of the patient (as represented by the MR scans) and the choice of treatment are the cause of ST. While previous related MR-based glioblastoma ST studies have focused only on the direct mapping of MR scans to ST, they have not included the underlying causal relationship between treatments and ST. To address this limitation, we propose a treatment-conditioned regression model for glioblastoma ST that incorporates treatment information in addition to MR scans. Our approach allows us to effectively utilize the data from all of the treatments in a unified manner, rather than having to train separate models for each of the treatments. Furthermore, treatment can be effectively injected into each convolutional layer through the adaptive instance normalization we employ. We evaluate our framework on the BraTS20 ST prediction task. Three treatment options are considered: Gross Total Resection (GTR), Subtotal Resection (STR), and no resection. The evaluation results demonstrate the effectiveness of injecting the treatment for estimating GBM survival.

\end{abstract}


\section{Introduction}

Glioblastoma (GBM) is one of the most aggressive malignant brain tumors, with a particularly poor survival rate. Despite significant advances in medical treatments for brain tumors, the median survival duration for GBM patients remains limited~\cite{kaur2022state}. The prognosis of GBM can vary widely depending on the patient's individual status and treatment approach.

We propose to develop a data-driven deep learning (DL) model to answer the question of whether a GBM patient would have a longer survival time (ST) with alternative treatments, based on the preoperative magnetic resonance (MR) scans. Our goal is to enable treatment-specific ST prediction, allowing for a comprehensive comparison of different treatments for a given patient. This capability is highly desired in the field of precision medicine, as it provides valuable insights for informed treatment planning.

The complementary status is a basis for an accurate prediction of ST. In addition to the tabular electronic health record (EHR)~\cite{lacroix2001multivariate}, there is increasing evidence that the multi-parametric MR scans have shown great potential for GBM prognosis~\cite{chang2016multimodal,nie2019multi}. Richer information can be expected from multimodal imaging. However, previous studies in MR-based GBM survival prediction have primarily focused on modeling a direct mapping from GBM patient status (represented as MR scans) to ST with direct models \cite{kaur2022state}, ignoring the underlying causal relationship between treatments and ST.
 
In practice, the influence of both the current state of the patient and the selection of treatment on the effect of the outcome is well justified \cite{athey2016recursive,shalit2017estimating}. However, the previous direct models either focused on a single treatment~\cite{chang2016multimodal,nie2019multi,chaddad2016quantitative} or did not incorporate the impact of the treatment~\cite{baid2020overall,huang2021overall}. In particular, without taking treatment into account, it is not possible to achieve a comparison between treatments to instruct the following treatment planning~\cite{tan2022multi}. Although a possible solution is to have different direct models for each treatment, there is only a limited amount of training data that can be used for each model.

To the best of our knowledge, this is the first attempt at a DL-based multimodal medical imaging study for individualized multi-treatment advisory, which takes both comprehensive multi-parametric preoperative MR status depictions and treatment selection into consideration for GBM survival inference. By considering the patient's MR scans and incorporating the treatment information in a unified framework, our proposed approach aims to unravel the intricate relationship between treatments and ST, thereby facilitating more accurate and personalized treatment decisions.
 
To this end, in this work, we propose to address the above issues with a treatment-conditioned GBM survival regression model using preoperative MR scans. Instead of training independent models for each treatment, we are able to use all treatment data in a unified manner. Specifically, we adopt a multichannel convolutional backbone as the previous direct model~\cite{huang2021overall} to aggregate the information in registered multiparametric MR scans and the corresponding segmentation map. Then, the treatment vector is integrated with either concatenation in the fully connected layer or the adaptive instance normalization (AdaIN) as in~\cite{liu2020auto3d,xu2019view,huang2018introvae,nguyen2019hologan,zakharov2019few} in each convolutional layer of the survival time prediction network. In particular, we normalize the mean ($\mu$) and variance ($\sigma$) of AdaIN layers to align with the relative-pose code $z$ instead of the feature map itself. Injecting treatment at earlier stages in a more interactive manner can potentially contribute to fine-grained treatment-conditioned modeling.   

We evaluate the performance of our framework on the BraTS2020 survival prediction task~\cite{bakas2018identifying}, considering three potential types of tumor resection. Our evaluation results demonstrate the effectiveness of our approach in estimating GBM survival and the benefit of the introduction of treatment.

\begin{figure}[t]
\begin{center}
\includegraphics[width=1\linewidth]{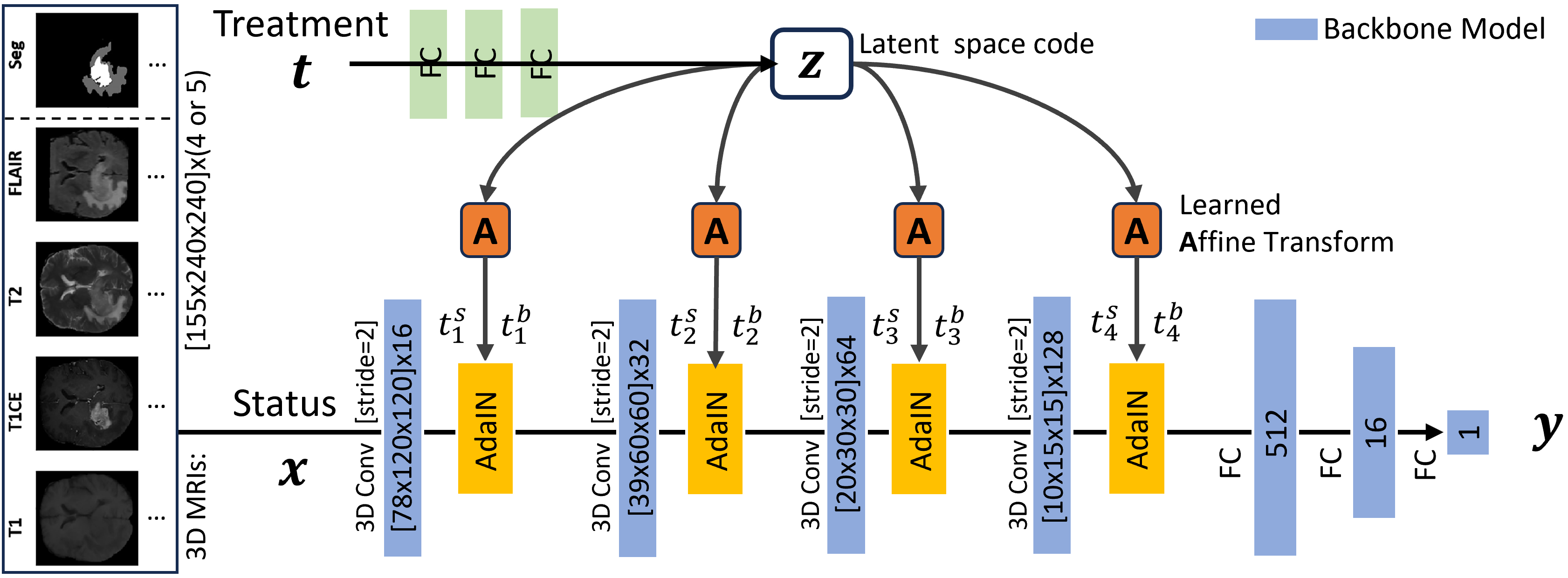}
\end{center}  
\caption{Illustration of our framework for the treatment-wise GBM survival time estimation with multi-parametric preoperative MR scans and segmentation mask. The treatment is injected by the AdaIN to every convolutional layer.} 
\label{ccc}\end{figure}

\section{METHODS}

The patient status $\bm{s}$ and treatment $\bm{t}$ are used to predict the effect $\bm{y}$ in individualized treatment effect inference. Specifically, in GBM survival analysis, multiparametric MR scans and the corresponding segmentation masks have been widely used to represent the status $\bm{s}$, which includes the location and radiomic features. In addition to the medical imaging status, the choice of treatment may also affect survival. For generality, we have a one-hot treatment vector $\bm{t}\in\mathbb{R}^N$ for $N$ possible treatment choices. Although it can be simplified to 0/1 for the binary treatment case, we aim to train a unified model $f:\{\bm{s},\bm{t}\}\rightarrow\bm{y}, \forall \bm{t}$. 

For a fair comparison, we follow the previous work~\cite{huang2021overall} and adopt the four layers of convolutional neural networks for 3D multi-modality information fusion. Then, we make use of three fully connected layers, each with a one-dimensional output unit, to predict the scalar value of survival days. Moreover, in the previous work~\cite{huang2021overall}, the patient age and manually crafted radiomics features were concatenated with the first fully connected layer representations, and then processed by either a neural network or conventional machine learning methods. However, in the present work, we focus solely on investigating the neural network approach and do not include the manually crafted features. Nevertheless, these predictive features can be easily incorporated if needed.

Instead, our objective is to investigate the impact of inducing treatment $\bm{t}$ to achieve a more accurate ST estimation by modeling fine-grained treatmentwise $p(\bm{y}|\bm{t})$ with a unified framework. Thus, a critical problem is how to seamlessly induce treatment $\bm{t}$ into the status feature extraction to model the conditional distribution. In previous conditional generative works \cite{karras2019style,liu2020auto3d}, it has been shown that simple concatenation in a fully connected layer can be less effective in combining two features, as the previous convolution layers are not well-informed, and hard to learn their correlation with limited subsequent parameters.

Therefore, following previous successful conditional modeling works\cite{karras2019style,liu2020auto3d}, we propose to adopt adaptive instance normalization (AdaIN) \cite{huang2017arbitrary} to induce treatment $\bm{t}$ in each convolutional layer. Specifically, we process $\bm{t}$ with three multi-layer perceptron (MLP) layers (all with 16-dim) to produce a 16-dim vector $\bm{z}\in\mathbb{R}^{16}$ in the intermediate latent space $\mathcal{Z}$, the learned affine transformations, i.e., the linear layers of each layer, then specialize $\bm{z}$ to scalars of treatment-wise scale and bias $\bm{t_i'}=(t^s_i,t^b_i)$ in the $i$-th convolutional layer to control the AdaIN operators. Note that the dimensionality of $t^s_i$ or $t^b_i$ should match the number of feature maps in that layer, which requires the dimension of the linear layer to be twice that of $t^s_i$ or $t^b_i$. Specifically, we have the following AdaIN operation in each layer:
\begin{equation}
AdaIN(\bm{x}_i, \bm{t'}) = t_i^s\frac{\bm{x}_i-\mu(\bm{x}_i)}{\sigma(\bm{x}_i)} + t_i^b,
\end{equation}
where $\bm{x}_i$ is the extracted feature map after the $i$-th convolutional layer. As a result, $\bm{x}_i$ is individually normalized and subsequently scaled and biased using the corresponding scalar components from $\bm{t_i'}=(t^s_i,t^b_i)$. By doing so, it injects a stronger inductive bias of $\bm{t}$ into the DL model. The model $f$ is detailed in Fig. 1, which is trained by the mean absolute error (MAE), i.e., $|f(\bm{x},\bm{t})-\bm{y}|$.

\begin{table}[t]
\centering
\caption{Numerical comparisons of different inputs and their combination methods. The best results are indicated in \textbf{bold}. The standard deviation is reported based on five random trials.} \vspace{+5pt}
\resizebox{0.8\linewidth}{!}{
\begin{tabular}{l|c|c|c}
\hline
Status &  Treatment &  &MAE (in days) \\\hline\hline

4 MR scans~\cite{huang2021overall} & - & - & 165.8$\pm$6.2\\\hline
4 MR scans [This paper] & $\surd$ & Concatenation & 140.2$\pm$9.0\\
4 MR scans [This paper] & $\surd$ & AdaIN & \textbf{132.7}$\pm$7.8\\\hline\hline

4 MR scans + Segmentation mask~\cite{huang2021overall}& -  & - & 124.4$\pm$5.3\\\hline

4 MR scans + Segmentation mask [This paper]& $\surd$ & Concatenation & 103.5$\pm$4.9 \\

4 MR scans + Segmentation mask [This paper]& $\surd$ & AdaIN & \textbf{98.2}$\pm$5.6\\\hline 

\end{tabular}
}
\label{tabel:1}
\end{table}


\section{RESULTS}
In the BraTS 2020 dataset~\cite{brats20,bakas2018identifying}, the preoperative T1-weighted (T1), T1-weighted with contrast enhancement (T1-CE), T2-weighted (T2), and fluid attenuation inversion recovery (FLAIR) MR scans and the segmentation mask are collected as the status. Survival days (5-1767d) are reported for 236 subjects. 119 were treated with Gross Total Resection (GTR), 10 with Subtotal Resection (STR), and 107 had no treatment (NA). We use the subject-independent 5-fold cross-evaluation on the 236 subjects, in which both training and test sets contain three treatments. Note that the split is different from the previous direct model~\cite{huang2021overall}, which does not consider the distribution of treatments. In addition, it is difficult to independently train a separate model for STR with very limited samples. The training was performed on an NVIDIA A100 GPU for 200 epochs.

We input status $\bm{x}=$\{multi-parametric MRs, segmentation\} and treatment $\bm{t}=$\{GTR, STR, NA\} to predict the survival day $\bm{y}$. We adopt their multi-channel 3D convolutional network as our backbone to process the four-channel 3D MR scans, which can be extended to incorporate the segmentation mask as five channels. Our framework was implemented using the PyTorch deep learning toolbox~\cite{paszke2017automatic}. For performance quantification, the MAE is reported in Table 1. We can see that adding the treatment to the survival estimation can significantly reduce MAE.  Furthermore, the AdaIN injection shows a superior capability to effectively combine the patient status and treatment compared to the simple concatenation method at the first fully connected layer.

\section{CONCLUSION}

In this work, we introduced a novel and accurate survival estimation model for GBM, using preoperative MRI data and segmentation maps to model the fine-grained treatment-dependent distribution. Our framework efficiently injects the treatment vector into the status representation extraction process. Compared with conventional direct models that do not consider treatment~\cite{huang2021overall}, or simply concatenate the treatment vector into the FC layers, our model demonstrates superior and effective performance. Moreover, our approach can serve as a simple add-on module to enhance more advanced backbones used in direct models, such as vision transformers. In future work, we will replace the manual segmentation map with the predictions of advanced DL-based tumor segmentation models \cite{liu2023memory,liu2021adapting,liu2022self,liu2022act,liu2023incremental} to establish a fully automated analysis pipeline.
  
\vspace{+5pt}
\acknowledgments      
 
This work is supported by R01CA165221 and P41EB022544.

\bibliography{main} 
\bibliographystyle{spiebib} 

\end{document}